\documentclass[preprint,12pt]{elsarticle}

\journal{ISPRS Journal of Photogrammetry and Remote Sensing}

\usepackage{graphicx}
\usepackage{amsmath}
\usepackage{amssymb}
\usepackage{textcomp}
\usepackage{algorithm}
\usepackage{algpseudocode}
\usepackage{booktabs}
\usepackage[inline]{enumitem}
\usepackage[font=footnotesize,skip=0pt]{caption}
\usepackage[colorlinks=true, linkcolor=blue, citecolor=black, urlcolor=cyan]{hyperref}
\usepackage{cleveref}

\usepackage{setspace}
\begin{document}

\begin{frontmatter}

\title{BIM-Discrepancy-Driven Active Sensing for Risk-Aware UAV-UGV Navigation}

\author[inst1]{Hesam Mojtahedi}
\ead{hmojtahedi@ucsd.edu} 

\author[inst2]{Reza Akhavian*}
\ead{rakhavian@sdsu.edu}

\cortext[cor1]{Corresponding author}

\address[inst1]{Department of Electrical and Computer Engineering, University of California San Diego, La Jolla, CA, USA}
\address[inst2]{Department of Civil, Construction, and Environmental Engineering, San Diego State University, San Diego, CA, USA}

\begin{abstract}
This paper presents a BIM-discrepancy-driven active sensing framework for cooperative navigation between unmanned aerial vehicles (UAVs) and unmanned ground vehicles (UGVs) in dynamic construction environments. Traditional navigation approaches rely on static Building Information Modeling (BIM) priors or limited onboard perception. In contrast, our framework continuously fuses real-time LiDAR data from aerial and ground robots with BIM priors to maintain an evolving 2D occupancy map. We quantify navigation safety through a unified corridor-risk metric integrating occupancy uncertainty, BIM-map discrepancy, and clearance. When risk exceeds safety thresholds, the UAV autonomously re-scans affected regions to reduce uncertainty and enable safe replanning. Validation in PX4-Gazebo simulation with Robotec GPU LiDAR demonstrates that risk-triggered re-scanning reduces mean corridor risk by 58\% and map entropy by 43\% compared to static BIM navigation, while maintaining clearance margins above 0.4 m. Compared to frontier-based exploration, our approach achieves similar uncertainty reduction in half the mission time. These results demonstrate that integrating BIM priors with risk-adaptive aerial sensing enables scalable, uncertainty-aware autonomy for construction robotics.

\end{abstract}


\end{frontmatter}

\label{sec:intro}
\section{Introduction}
Construction sites are among the most dynamic, unstructured, and safety-critical environments for autonomous robots. 
Unlike factory floors or structured indoor spaces, these environments are marked by continual change. 
New buildings are erected, materials are relocated, and the movement of heavy machinery and workers can be unpredictable. 
Such conditions make autonomous navigation particularly challenging. 
Construction~4.0~\cite{forecael2020}, emphasizing automation and digitalization, is moving robotics from trial phases to regular use on construction sites. 
Recent systematic reviews indicate that robotics technologies are not only becoming more prevalent but are also starting to impact core construction processes by advancing perception, mobility, manipulation, and collaboration~\cite{liu2024construction, sofiat2021, emaminejad2022trustworthy}. 
These reviews further highlight that construction robotics is at a pivotal point: research activity has increased steadily since 2016, with advancements in sensing and autonomy, but large-scale, fully coordinated deployments remain limited.

Environmental sensing forms the foundation of autonomy in construction robotics. 
Perception systems such as vision, thermal imaging, LiDAR, and multimodal sensing enable robots to track dynamic objects, detect hazards, and respond effectively in uncertain environments~\cite{OKONKWO2025106075}. 
Construction sites are especially dynamic: layouts change daily, heavy equipment often operates near workers, and occlusions are common. 
Building Information Modeling (BIM) provides valuable static priors about anticipated site conditions. However, BIM cannot fully capture the evolving state of construction sites ~\cite{meyer2023geometric, amani2025safe}. 
Real-time sensing and adaptive algorithms must therefore be integrated with navigation and task planning to ensure safety and task performance ~\cite{rao2022real}.

A variety of sensing modalities have been employed previously, each with distinct advantages and limitations. 
LiDAR offers centimeter-level accuracy for 3D mapping but is susceptible to interference from dust, debris, and reflective surfaces, and incurs high costs~\cite{puri2020bridge}. 
Radar is capable of penetrating weather and visibility barriers to provide velocity data, though its precision is limited. 
Ultrasonic sensors are low-cost and energy-efficient but offer only a short and imprecise range. 
Vision sensors, including monocular, stereo, and RGB-D, supply rich semantic and texture information, yet are challenged by occlusions and variable lighting. 
Inertial and odometry systems provide motion estimates but are prone to drift when not fused with other sensing modalities. 
No single modality is sufficient on its own, making intelligent multi-sensor fusion essential for robust operation~\cite{tavassoli2023expanding}.

Equally important are the algorithms that process this sensor data. 
Simultaneous Localization and Mapping (SLAM) is widely used for robots to build maps while localizing themselves, but it generally assumes static environments. 
Vision-based convolutional neural network (CNN) detectors, such as YOLO, can identify objects of interest (e.g., scaffolding or workers), but similar-looking objects are difficult to distinguish in cluttered or occluded scenes, and generalization to new environments is limited. 
Classical path planning algorithms (e.g., A*, D*, and Rapidly-Exploring Random Tree (RRT)) must be adapted to navigate dynamic obstacles or to replan entirely. 
Probabilistic filters, such as Kalman and particle filters, address uncertainty but often face challenges with scalability and nonlinearity in large, dynamic environments.

Single-agent robotic systems offer distinctive benefits. 
Unmanned Aerial Vehicles (UAVs) provide rapid and comprehensive situational awareness for site mapping, change detection, and the inspection of otherwise inaccessible structures. 
UAVs deliver a global perspective that ground robots cannot match. 
In contrast, Unmanned Ground Vehicles (UGVs) offer persistence, higher payload capacity, and direct interaction with the environment for inspection, manipulation, and transportation. 
However, UAVs lack persistence and the ability to interact directly with the ground, while UGVs are limited in their spatial coverage and resolution for mapping large or intricate structural elements.

This contrast presents a valuable research opportunity, as there are currently no established frameworks for aerial–ground coordination in construction. 
Multi-robot systems have been explored for search and rescue~\cite{shahar2025ugvuav, munasinghe2024comprehensive}, agriculture, and defense~\cite{munasinghe2024comprehensive}; however, applications directly relevant to construction remain underexplored. 
Coordinated UAV–UGV coverage can provide both micro- and macro-level situational awareness, supporting site monitoring and inspection while addressing navigation challenges in dynamic and hazardous environments. 
A UAV–UGV system for mapping and navigation can leverage BIM priors in combination with real-time perception, wherein UAVs survey regions where as-built conditions diverge from BIM predictions and relay this information to UGVs for safer and more efficient navigation.

Despite recent advancements, several practical gaps remain in deploying autonomous systems on active construction sites. 
Many studies rely on a limited set of sensor modalities, short-duration experiments, or overly simplified assumptions about environmental conditions. 
In reality, construction sites require robust sensor fusion that can handle glare, dust, variable lighting, and frequent occlusions. 
Real-time performance is equally critical to balance mapping accuracy against computational limits on embedded hardware. 
Furthermore, long-term operations must address cumulative drift and maintain consistency across changing site states. 
Finally, a principled framework is needed to reconcile BIM-derived maps with evolving as-built conditions and to harmonize human intentions with the dynamic final state of construction environments.

\textbf{Contributions.}
This paper presents a \emph{BIM-discrepancy-driven active sensing framework} designed for UAV and UGV teams operating within dynamic construction environments. 
Distinct from conventional local planning methods, which often assume static site conditions or depend exclusively on onboard perception, the proposed approach integrates BIM priors with real-time LiDAR observations. 
This integration supports the maintenance of an evolving, risk-aware representation of the construction site. 
The principal contributions of this work are as follows:
\begin{itemize}
    \item \textbf{BIM-initialized probabilistic mapping:} Introduction of a two-dimensional occupancy grid, initialized using BIMs and continuously refined through log-odds fusion of LiDAR data. This allows for geospatial alignment between design and as-built conditions.
    \item \textbf{Risk-driven UAV coordination:} A unified corridor risk metric that integrates occupancy uncertainty, BIM–map discrepancy, and clearance. This metric triggers UAV re-scanning when the UGV encounters regions deemed unsafe or occluded, thereby enhancing operational safety and efficiency.
    \item \textbf{Experimental validation:} Comprehensive evaluation in a PX4–Gazebo simulation environment, utilizing Robotec GPU LiDAR and ROS~2 integration. The results demonstrate the accurate detection of structural deviations, a reduction in uncertainty, and improved navigation safety for UGVs in dynamic, BIM-derived construction environments.
\end{itemize}

Taken together, these gaps motivate a framework that uses BIM as a soft prior, measures where reality diverges from design, and triggers targeted aerial re-sensing only when needed to keep a ground robot safe and efficient.

\label{sec:related}
\section{Related Works}

\subsection{Active Sensing and Frontier-Based Exploration}
Active perception allows robots to adapt their sensing strategies based on the anticipated informational value of new observations. 
This approach enables more efficient and targeted data collection in complex environments. 
Traditional exploration strategies often rely on the concept of \emph{frontiers}, which are defined as the boundaries separating known and unknown areas within an environment. 
Yamauchi~\cite{yamauchi1997frontier} first introduced frontier-based exploration for 2D occupancy mapping, which was later extended by multi-robot clustering~\cite{solanas2004coordinated, faigl2013determination}, cost-based gain metrics~\cite{joho2007autonomous}, and Voronoi-based decentralized coordination~\cite{wu2007voronoi, simmons2000coordination, burgard2005coordinated}. 
Although frontier-based planners are efficient, they typically assume static environments and treat all unexplored regions equally. 
This uniform treatment can reduce their effectiveness in dynamic and structured settings, such as construction sites, where environmental conditions change frequently. 
Next-Best-View (NBV) and Receding Horizon Next-Best-View (RH-NBV) approaches extend frontier concepts to 3D volumetric mapping by optimizing viewpoint selection to maximize information gain~\cite{bircher2016receding, papachristos2018autonomous, charrow2015information}. 
While these methods are effective for detailed exploration tasks, they are computationally intensive and frequently require the use of dense Truncated Signed Distance Field (TSDF) or voxel-based maps. 
By contrast, the approach presented in this work emphasizes \emph{risk-triggered sensing} guided by discrepancies and uncertainties identified through BIM. 
This strategy enables scalable and interpretable mapping in semi-structured, evolving environments without the need for continuous NBV optimization.

\subsection{Active SLAM and Planning Under Uncertainty}
Active Simultaneous Localization and Mapping (ASLAM) integrates motion planning with uncertainty reduction to balance exploration and localization accuracy~\cite{bry2011rapidly, agha2014slap, carlone2014active, bourgault2002information, carrillo2018autonomous}. 
Information-theoretic methods formulate objectives in terms of entropy reduction or Kullback–Leibler divergence, while belief-space planners such as Rapidly-Exploring Random Belief Trees (RRT-Bel)~\cite{bry2011rapidly} explicitly reason over uncertainty. 
Although these approaches are capable of generating consistent maps in previously unknown environments, they often operate under the assumption that prior structural information is not available. 
In contrast, the proposed method leverages BIM as a probabilistic prior, measuring isk and uncertainty relative to known structures instead of solely unobserved areas. 
This shift from pure exploration to \emph{discrepancy-aware re-exploration} better suits the dynamic and partially known nature of construction sites.

\subsection{3D Mapping and Occupancy Representations}
Modern mapping frameworks such as OctoMap~\cite{hornung2013octomap} and Voxblox~\cite{oleynikova2017voxblox} represent occupancy and Euclidean Signed Distance Fields (ESDFs) in three dimensions, enabling gradient-based planning and volumetric reconstruction. 
These probabilistic methods extend classical log-odds occupancy models~\cite{bongard2008probabilistic} to 3D, supporting continuous updates from range data. 
However, construction robots are generally deployed on relatively planar sites or within corridors, where comprehensive volumetric mapping can be redundant and unnecessarily computationally demanding. 
Consequently, the framework described in this paper utilizes a 2D occupancy grid representation at $z=0$, which serves as a practical simplification for ground-level navigation. 
This design choice also preserves compatibility with future extensions to 2.5D or ESDF models for vertical risk assessment.

\subsection{UAV–UGV Cooperative Mapping}
Heterogeneous aerial–ground cooperation exploits the complementary sensing capabilities of UAVs and UGVs: UAVs offer wide-area coverage and global visibility, while UGVs provide detailed, persistent mapping at ground level~\cite{maboudi2023review}. 
Prior work has demonstrated coordinated exploration using frontier allocation~\cite{corah2019communication, simmons2000coordination}, Voronoi partitioning~\cite{wu2007voronoi}, and communication-aware planning~\cite{baca2021autonomous, spurny2019cooperative, kashyap2025simulations}. 
Typically, these methods aim to achieve uniform coverage or complete exploration rather than targeted information acquisition relevant to specific operational needs. 
In contrast, the proposed framework introduces a BIM-informed risk trigger, whereby the UAV is deployed only when the UGV’s local corridor displays high uncertainty, reduced clearance, or significant BIM–map discrepancies. 
This mechanism ensures that aerial interventions are both selective and efficient.

\subsection{BIM Integration and Prior-Guided Mapping}
BIMs encode both geometric and semantic representations of designed environments and have been extensively investigated for applications in as-built verification, progress monitoring, and robotic localization. 
Early work by Bosché and Haas~\cite{bosche2009automated} introduced automated retrieval of 3D CAD model objects from construction range images, establishing one of the first robust pipelines for scan-to-BIM recognition and as-built verification. 

Turkan et~al.~\cite{turkan2012automated} extended this approach by fusing 3D laser scan data with 4D BIM schedules for automated progress tracking, demonstrating how time-stamped as-built models can support dynamic construction monitoring. 
Building on these foundations, Golparvar-Fard et al.~\cite{golparvar2011vision} presented a computer-vision framework that reconstructs construction scenes from unordered site photographs via structure-from-motion and multi-view stereo, fusing them with BIM geometry to automatically detect deviations under severe occlusions and lighting variations. 
More recently, Pal et al.~\cite{PAL2023100247} advanced BIM-based monitoring toward Digital Twin (DT) integration, enabling bidirectional data exchange between physical and digital assets for predictive, AI-driven construction control. 
Despite these advances, the majority of prior work focuses on offline or periodic model comparison rather than facilitating dynamic, uncertainty-aware map updates necessary for real-time autonomous robotic operations.

The method presented here extends BIM integration beyond passive alignment to encompass active risk assessment. 
In this framework, the BIM serves as a soft geometric prior that evolves online as the occupancy map is incrementally updated through UAV and UGV sensing. 
By quantifying both BIM–map discrepancies and occupancy uncertainty, the system is able to actively identify deviations from the intended design within the built environment and direct sensing resources to these areas as needed.

\subsection{Identified Gaps and Novelty}
Unlike previous active SLAM, NBV, or BIM-based mapping frameworks, our approach uses geospatial LiDAR data for real-time, uncertainty-aware updates, integrating BIM–map discrepancy, occupancy uncertainty, and clearance into a unified \emph{corridor risk metric}. 
This metric determines the timing and location of UAV re-scanning, facilitating safety-conscious and resource-efficient map updates that explicitly account for the ever-changing conditions present in construction environments.

\label{sec:overview}
\section{Overview of the Developed Framework}
\begin{figure}[hbt!]
    \centering
    \includegraphics[width=1.1\linewidth]{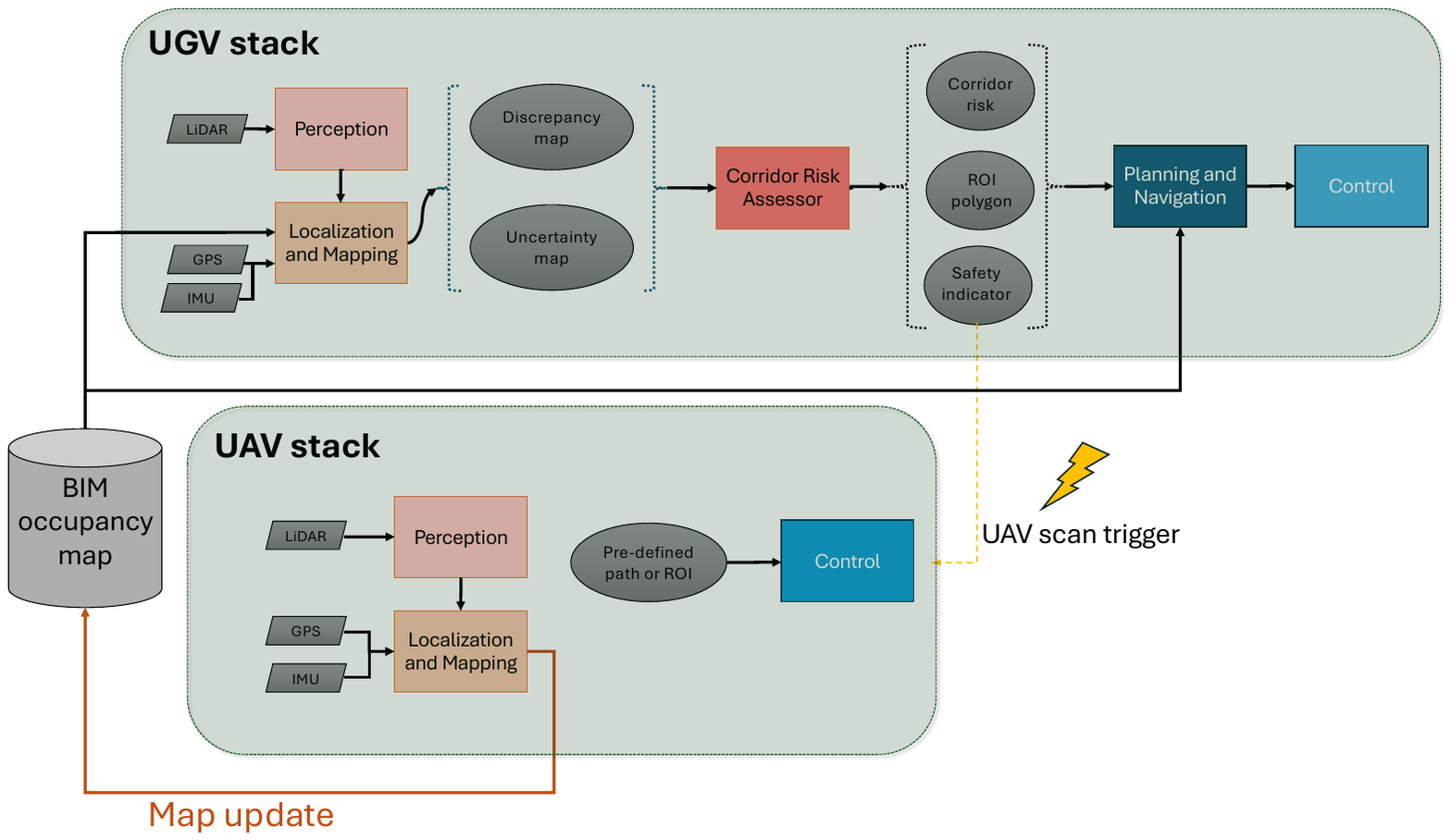}
    \caption{
        System architecture of the proposed UAV–UGV autonomous navigation framework. 
    }
    \label{fig:system_architecture}
\end{figure}

The proposed framework utilizes the BIM as a probabilistic geometric prior, encoding the anticipated layout of the job site and iteratively refining this model with real-time LiDAR data. 
By integrating BIM-based priors with real-time perceptual data, the system establishes a link between design-phase information and actual site conditions. 
This integration enables adaptive navigation in environments that are only partially known and subject to ongoing change, without necessitating complete three-dimensional reconstruction or exhaustive exploration. Figure \ref{fig:system_architecture} illustrates the system architecture of the developed framework, showing separate processing stacks for the UAV and UGV.

\subsection{System Concept}
At initialization, the BIM is rasterized into a 2D probabilistic occupancy grid representing as-designed free and occupied regions. 
Both the UAV and UGV are deployed in a PX4–Gazebo simulation environment and are equipped with GPU-accelerated LiDAR sensors to ensure high-fidelity data acquisition. 
Incoming \texttt{LaserScan} data from both platforms are systematically fused with the BIM prior using a log-odds update model. 
This process generates an evolving as-built map that captures real-world deviations, such as newly introduced barriers, removed walls, or displaced materials. 
Each cell in the occupancy grid maintains a probabilistic estimate of occupancy, which is dynamically updated with each new range measurement. 
This approach ensures that the map accurately reflects the current state of the construction site.

\subsection{Corridor-Based Risk Assessment}
The UGV follows a globally optimized trajectory derived from the BIM prior and simultaneously monitors a safety corridor surrounding its path at the local level. 
This safety corridor delineates the region of immediate operational concern and forms the basis for subsequent risk assessment. 
Three complementary layers are continuously analyzed in real time:
\begin{itemize}
    \item \textbf{Clearance:} calculated as the Euclidean distance to the nearest occupied cell, providing an estimate of available traversable space.
    \item \textbf{Uncertainty:} determined by calculating the Shannon entropy of occupancy probabilities, which quantifies the level of confidence in current map estimates.
    \item \textbf{BIM–map discrepancy:} measured as the absolute difference between BIM prior probability and observed occupancy probability, thereby identifying structural deviations between the design and as-built environment.
\end{itemize}
These three layers are integrated into a single scalar \emph{corridor risk score}, which quantitatively assesses navigation safety within the UGV’s immediate look-ahead region. 
When the risk score remains below a predefined threshold, the UGV proceeds with its mission. 
If the score surpasses this threshold, additional sensing is initiated before navigation continues.

\subsection{Risk-Triggered UAV Re-Scanning}
When high-risk regions are detected, the UAV is autonomously deployed to execute a brief, predefined coverage scan focused on the affected corridor segment. 
This predefined scan pattern, such as a lateral sweep or circular pass, enables efficient data collection without the computational overhead of next-best-view optimization. 
During the scan, the UAV operates \texttt{slam\_toolbox} in online mode to generate an updated two-dimensional occupancy grid from its \texttt{LaserScan} observations. 
The resulting map is then integrated into the global occupancy grid, thereby reducing uncertainty and rectifying any BIM–map discrepancies. 
Once the updated map meets the required clearance and risk thresholds, the UGV replans its trajectory and resumes operation in a safe manner.

\subsection{Summary of Workflow}
The proposed framework establishes a closed-loop system that integrates BIM priors, LiDAR-based perception, and risk-triggered aerial updates through a cyclical perception–update–planning process:
\begin{enumerate}
    \item Initialize the occupancy grid from BIM geometry.
    \item Fuse incoming LiDAR data from both UAV and UGV into the shared map.
    \item Continuously evaluate clearance, uncertainty, and BIM–map discrepancy along the UGV’s corridor.
    \item Trigger UAV scanning when the corridor risk exceeds a safety threshold.
    \item Fuse the UAV’s updated occupancy grid, re-evaluate risk, and replan if necessary.
\end{enumerate}
This coordinated pipeline ensures that an up-to-date representation of the evolving construction site is maintained, enabling the UGV to navigate with enhanced safety and spatial awareness while minimizing unnecessary UAV deployments.

\label{sec:method}
\section{Methodology}
\label{sec:methodology}

We propose a BIM-discrepancy-driven active sensing framework for cooperative UAV–UGV navigation on construction sites. 
The BIM acts as a \emph{soft geometric prior} for global planning, while targeted aerial sensing is dispatched only where the BIM and the online map disagree within the UGV’s forward corridor. 
The pipeline has four stages: (1) BIM-initialized mapping, (2) corridor risk assessment and trigger, (3) ROI extraction, and (4) predefined UAV scan with map fusion.

\subsection{World and Mapping}
\label{sec:world_mapping}

\paragraph{2D grid and layers}
We represent the site at ground level ($z=0$) as a 2D occupancy grid with resolution~$\Delta$. 
The BIM prior~$\tilde{\mathcal M}$ encodes as-designed occupancy; the online map~$\mathcal M_t$ estimates as-built occupancy.
Formally, the BIM prior is given by
\[
\tilde{\mathcal M} = \{\tilde{O}(i,j)\}_{i,j},
\]
where $\tilde{O}(i,j) \in \{0,1\}$ is the BIM-derived binary occupancy (1 = occupied, 0 = free), and the online map at time~$t$ is
\[
\mathcal M_t = \{O_t(i,j)\}_{i,j},
\]
where $O_t(i,j) \in [0,1]$ denotes the estimated occupancy probability.

\paragraph{Scope and 2D assumption}
We model the site at ground level ($z=0$) with a 2D occupancy grid. 
This choice reflects our target use case of \emph{single-level interior corridors}, where vertical variability is low (no mezzanines within the UGV corridor), and the UGV’s safety margin is dominated by lateral clearance. 
The onboard \texttt{LaserScan} (planar) and the UAV’s low-altitude planar sweeps provide predominantly 2D range observations; thus, collapsing to 2D avoids spurious “ghost” occupancy from partial vertical returns and keeps the perception–planning loop real-time on embedded hardware. 

\paragraph{Motivation for a probabilistic formulation}
To effectively combine BIM priors with real-time LiDAR measurements, we need a map representation that can
\begin{enumerate*}
    \item explicitly represents uncertainty,
    \item handle conflicts between design-time structures and as-built observations, and
    \item supports fast online updates for real-time navigation.
    A probabilistic occupancy approach meets all three requirements: it models each grid cell as a Bernoulli variable, provides interpretable probabilities for entropy-based uncertainty metrics, and naturally handles disagreements with BIM priors. Among probabilistic methods, the log-odds representation is beneficial because it converts Bayesian updates into simple additive operations, enabling high-rate map fusion without intensive computation.
\end{enumerate*}

\paragraph{Bayesian occupancy mapping}
To incrementally update occupancy estimates as new sensor measurements arrive, we employ a recursive Bayesian formulation in log-odds space. This representation offers computational efficiency and numerical stability compared to directly maintaining probability values, as it converts probability multiplication into addition and avoids numerical underflow.

\paragraph{Log-odds fusion with BIM prior}
Each grid cell $(i,j)$ maintains a log-odds occupancy value $L_t(i,j)$ that accumulates evidence from all range measurements up to time~$t$. 
Given the incoming sensor observations~$\mathcal{Z}_t$, the map update follows the standard recursive Bayes formulation in log-odds form:
\begin{align}
L_t(i,j)
&= 
\operatorname{clip}\!\left(
L_{t-1}(i,j)
+ \sum_{z \in \mathcal{Z}_t} \ell(i,j;z),\;
L_{\min},\,L_{\max}
\right), \\[4pt]
O_t(i,j) &= \frac{1}{1 + \exp[-L_t(i,j)]}.
\end{align}
Here, $\ell(i,j;z)$ denotes the \emph{inverse sensor model}, representing the additive evidence contributed by a single LiDAR beam~$z$:
\begin{equation}
\ell(i,j;z) =
\begin{cases}
\ln\!\dfrac{P_{\text{occ}}}{1 - P_{\text{occ}}}, & \text{if the beam hits } (i,j),\\[10pt]
\ln\!\dfrac{P_{\text{free}}}{1 - P_{\text{free}}}, & \text{if the beam does not hit } (i,j).
\end{cases}
\label{eq:inverse_sensor_model}
\end{equation}

Typical parameters are $P_{\text{occ}}=0.9$ and $P_{\text{free}}=0.35$, yielding log-odds increments of approximately $+2.197$ and $-0.619$, respectively. 
The clipping function 
$\operatorname{clip}(x,a,b)=\min(\max(x,a),b)$ 
constrains the log-odds within $[L_{\min},L_{\max}]$ (usually $[-5,5]$), preventing numerical overflow and overconfidence due to repeated consistent updates. 
The corresponding occupancy probability $O_t(i,j)\in[0,1]$ expresses the current belief that cell $(i,j)$ is occupied.

\medskip
\noindent
We initialize the map from a \emph{soft BIM prior}:
\begin{align}
L_0(i,j) &= \ln \frac{\pi_0(i,j)}{1 - \pi_0(i,j)}, \\
\pi_0(i,j) &=
\begin{cases}
p_{\text{occ}}, & \tilde{O}(i,j) = 1,\\[3pt]
1 - p_{\text{occ}}, & \tilde{O}(i,j) = 0,
\end{cases}
\end{align}
where $p_{\text{occ}} \in [0.6, 0.9]$ represents BIM’s confidence in occupied regions, and $\tilde{O}(i,j)$ is the BIM-derived binary occupancy defined above.

\paragraph{Uncertainty and discrepancy layers.}
For each grid cell $(i,j)$, two auxiliary quantities are derived from the occupancy probability $O_t(i,j)$ to guide active sensing: 
the \emph{uncertainty layer} $H_t(i,j)$ and the \emph{BIM–map discrepancy layer} $D_t(i,j)$.

\medskip
\noindent
\textbf{Uncertainty} 
The Shannon entropy of the occupancy probability quantifies the local uncertainty:
\begin{align}
H_t(i,j)
&= -\,O_t(i,j)\,\log O_t(i,j)
- \big(1 - O_t(i,j)\big)\,\log\!\big(1 - O_t(i,j)\big).
\end{align}
High entropy ($H_t \approx \log 2$) occurs when $O_t(i,j) \approx 0.5$, indicating ambiguity between free and occupied space, 
while low entropy ($H_t \approx 0$) reflects high confidence in either state.

\medskip
\noindent
\textbf{BIM discrepancy} 
To measure disagreement between the as-designed BIM prior $\tilde{O}(i,j)$ and the as-built map $O_t(i,j)$, we define:
\begin{equation}
D_t(i,j) = \big|O_t(i,j) - \tilde{O}(i,j)\big|.
\end{equation}
Large discrepancies ($D_t(i,j) > 0.5$) signify locations where the current sensor observations contradict the BIM model:
\begin{itemize}
    \item $\tilde{O}(i,j) = 0$ and $O_t(i,j) \approx 1$: newly appearing obstacles or temporary structures.
    \item $\tilde{O}(i,j) = 1$ and $O_t(i,j) \approx 0$: removed or relocated elements.
\end{itemize}
Together, $H_t(i,j)$ captures \emph{how uncertain} the map is, while $D_t(i,j)$ captures \emph{where BIM and reality diverge}.

\paragraph{Clearance via Euclidean Distance Transform (EDT)}
To evaluate traversability, the occupied set at time~$t$ is defined as
\[
\mathcal{O}_t = \{\mathbf{x} \mid O_t(\mathbf{x}) > \tau_{\text{occ}}\},
\]
where $\tau_{\text{occ}}$ is the occupancy threshold for classifying cells as obstacles.  
The Euclidean Distance Transform (EDT) $\mathrm{DT}_t(\cdot)$ assigns to each grid location the shortest Euclidean distance to the nearest occupied cell:
\begin{equation}
\mathrm{DT}_t(\mathbf{x}) =
\min_{\mathbf{y} \in \mathcal{O}_t} \|\, \mathbf{x} - \mathbf{y} \,\|_2,
\label{eq:edt}
\end{equation}
The instantaneous UGV clearance margin at position $\mathbf{x}$ is then defined as
\begin{equation}
\mathrm{clr}_t(\mathbf{x}) = \mathrm{DT}_t(\mathbf{x}) - r_{\text{ugv}},
\end{equation}
where $r_{\text{ugv}}$ is the UGV’s effective radius.  
A positive margin $\mathrm{clr}_t(\mathbf{x})>0$ indicates safe passage, 
whereas $\mathrm{clr}_t(\mathbf{x}) \le 0$ denotes potential collision or insufficient clearance.

\subsection{Corridor Risk and UAV Triggering}
\label{sec:corridor_trigger}

A nominal UGV path~$\mathcal{P}$ is first planned on the BIM-derived prior map (e.g., using RRT* search), 
with obstacle inflation radius $r_{\text{ugv}} + m_{\min}$ to ensure safety margins.  
A \emph{safety corridor} of half-width~$d_c$ is defined around~$\mathcal{P}$, representing the spatial band used for risk assessment.

Within a forward look-ahead window~$\mathcal{W}_\ell(t)$ of arc length~$\ell$, 
a per-cell \emph{risk score} is computed by combining occupancy uncertainty and BIM–map discrepancy:
\begin{equation}
\begin{split}
r_t(i,j) &= \alpha\,H_t(i,j) + \beta\,D_t(i,j),\\[3pt]
R_\ell(t) &= 
\frac{1}{|\Omega_t|}
\sum_{(i,j)\in \Omega_t} r_t(i,j),
\end{split}
\label{func:risk}
\end{equation}
where $\Omega_t = \mathcal{W}_\ell(t)\cap\mathcal{C}$ denotes the subset of the forward window that lies inside the safety corridor~$\mathcal{C}$.
Where $\alpha$ and $\beta$ are weighting coefficients controlling the contribution of uncertainty and discrepancy, respectively.  
The aggregated corridor risk~$R_\ell(t)$ thus quantifies the average mapping confidence along the near-future segment of the UGV path. In all experiments, we set $\alpha = \beta = 0.5$, assigning equal weight to uncertainty and BIM–map discrepancy.

\medskip
\noindent
We compute a scalar corridor risk from uncertainty and BIM–map discrepancy, and enforce clearance via a separate safety constraint. A UAV rescan is triggered whenever the forward corridor is deemed unsafe—either due to elevated risk or insufficient clearance:
\begin{equation}
R_\ell(t) > \tau_{\text{safe}}
\quad \text{or} \quad
\min_{\mathbf{p}_k \in \mathcal{W}_\ell(t)}
\mathrm{clr}_t(\mathbf{p}_k) < m_{\min}.
\end{equation}
The first condition corresponds to high uncertainty or strong BIM–map disagreement, 
while the second detects potential physical collisions.  
This dual criterion ensures that aerial sensing is invoked only when the UGV’s local environment cannot be safely inferred from onboard perception alone.

\subsection{Region of Interest (ROI)}
\label{sec:roi}

To focus aerial sensing on the most critical areas, we extract a high-risk Region of Interest (ROI) within the forward look-ahead window~$\mathcal{W}_\ell(t)$. Rather than re-scanning the entire corridor, we identify the contiguous subset of $N_{\text{roi}}$ cells exhibiting the highest average risk. Formally, the ROI $\mathcal{R}^\star$ is defined as:
\begin{equation}
\mathcal{R}^\star
= 
\arg\max_{\substack{\mathcal{R} \subset \mathcal{W}_\ell(t) \\ |\mathcal{R}| = N_{\text{roi}}}}
\frac{1}{N_{\text{roi}}}
\sum_{(i,j) \in \mathcal{R}} r_t(i,j).
\end{equation}
This sliding-window search efficiently locates the spatial cluster of maximum combined uncertainty and BIM--map discrepancy.  
Its computational complexity is linear in the window size, i.e., $O(|\mathcal{W}_\ell|)$, enabling real-time operation on embedded systems.

\subsection{UAV Scanning and BIM–Map Update (via SLAM Toolbox)}
\label{sec:uav_scan}

When the UGV detects elevated corridor risk (Sec.~\ref{sec:corridor_trigger}), it halts navigation and triggers a UAV scan mission. 
The UAV performs a predefined coverage flight over the surrounding area at altitude~$z_0$, collecting 2D \texttt{LaserScan} data while running \texttt{slam\_toolbox} in mapping mode. 
This process builds an updated 2D occupancy grid that captures the current as-built layout of the construction site, reducing the effects of occlusions and limited ground-level visibility.

\paragraph{Map update.}
Upon completion of the flight, the UAV exports the generated occupancy map using the \texttt{/save\_map} service and republishes it as the updated site map~$\mathcal{M}_t$. 
The new map replaces the previous UGV map entirely, ensuring both robots operate within a consistent global frame. 
This global refresh allows the system to incorporate structural or environmental changes that may have occurred since the last scan. In all experiments, we used a full-map replace after each UAV sweep; ROI paste is supported but left for future evaluation.

\paragraph{BIM comparison.}
The refreshed occupancy grid is compared against the BIM prior~$\tilde{\mathcal{M}}$ to update the previously defined discrepancy layer $D_t(i,j)$ (see Sec.~\ref{sec:world_mapping}). 
This highlights deviations such as missing walls, temporary obstructions, or newly added materials.

\paragraph{Replanning.}
Following the map update, the UGV replans its path~$\mathcal{P}$ on~$\mathcal{M}_t$ using the same global planner (e.g., A* or RRT* with inflation~$r_{\text{ugv}} + m_{\min}$). 
Once a safe path is obtained, navigation resumes toward the goal. 
This cycle of risk detection, UAV-assisted mapping, and replanning continues until the UGV successfully reaches its destination.
Algorithm~\ref{alg:perception_update} summarizes the full perception–update loop, including risk evaluation, UAV triggering, ROI extraction, and map integration.

\begin{algorithm}[H]
\caption{Perception–update loop with risk-triggered UAV re-scan}
\label{alg:perception_update}
\scriptsize
\begin{algorithmic}[l]
\Require BIM prior $\tilde{\mathcal M}$, start $A$, goal $B$, corridor half-width $d_c$, look-ahead arc $\ell$, thresholds $\tau_{\text{safe}},\, m_{\min}$
\Ensure Safe UGV navigation with on-demand UAV re-scan and 2D map update
\vspace{3pt}
\State $\mathcal{M}_0 \gets \tilde{\mathcal M}$ \Comment{soft BIM prior rasterized to grid}
\State $\mathcal{P} \gets \textsc{PlanPath}(A,B;\mathcal{M}_0)$;\quad $\mathcal{C} \gets \textsc{Corridor}(\mathcal{P}, d_c)$
\While{UGV not at $B$}
  \State $\mathcal{M}_t \gets \textsc{FuseUGVScan}(\mathcal{M}_{t-1}, \texttt{LaserScan})$
  \State $H_t, D_t, \mathrm{clr}_t \gets \textsc{UpdateLayers}(\mathcal{M}_t, \tilde{\mathcal M})$
  \State $\mathcal{W}_\ell(t) \gets \textsc{ForwardWindow}(\mathcal{P}, \ell, \text{pose at }t)$
  \State $R_\ell(t) \gets \frac{1}{|\Omega_t|}\sum_{(i,j)\in \Omega_t}\big(\alpha H_t(i,j)+\beta D_t(i,j)\big)$
  \State \hskip1.3em where $\Omega_t \!=\! \big(\mathcal{W}_\ell(t)\cap \mathcal{C}\big)$
  \State $c_{\min} \gets \min_{\mathbf{p}\in \mathcal{W}_\ell(t)} \mathrm{clr}_t(\mathbf{p})$
  \If{$R_\ell(t) \le \tau_{\text{safe}}$ \textbf{and} $c_{\min} \ge m_{\min}$}
     \State \textbf{continue} along $\mathcal{P}$
  \Else
     \State $\mathcal{R}^\star \gets \textsc{RoiMaxFilter}\!\left(\{\alpha H_t+\beta D_t\}\ \text{restricted to } \Omega_t\right)$
     \State \textsc{CommandUavScan}\ $\Gamma(\mathcal{R}^\star, z_0)$
     \State $\widehat{\mathcal{M}} \gets \textsc{SlamToolboxMap}()$ \Comment{updated \texttt{/map} from UAV sweep}
     \State $\mathcal{M}_t \gets \textsc{MapHandoff}(\mathcal{M}_t, \widehat{\mathcal{M}})$ \Comment{full-map replace (ROI paste supported but unused)}
     \State $H_t, D_t, \mathrm{clr}_t \gets \textsc{UpdateLayers}(\mathcal{M}_t, \tilde{\mathcal M})$
     \State $\mathcal{P} \gets \textsc{PlanPath}(A,B;\mathcal{M}_t)$ \Comment{inflate by $r_{\text{ugv}} + m_{\min}$}
  \EndIf
\EndWhile
\end{algorithmic}
\end{algorithm}

\section{System Setup}
\label{sec:system_setup}

The simulation framework integrates PX4, Gazebo Harmonic, RGL, and ROS~2 into a unified environment in which a UAV and a UGV operate cooperatively within a BIM-referenced environment.
This architecture facilitates risk-aware navigation and active mapping, supporting safe and adaptive operation in dynamic, built environments.

\subsection{Simulation and Vehicle Configuration}
Both robotic platforms are simulated in Gazebo Harmonic using physically based dynamics, accurate inertia parameters, and sensor models.
The UAV is equipped with a downward-facing GPU LiDAR and an onboard inertial measurement unit (IMU), whereas the UGV is outfitted with a 2D planar LiDAR mounted 0.4 $\mathrm{m}$ above ground level for horizontal scanning.

Each vehicle operates the PX4 Autopilot in software-in-the-loop (SITL) mode, connecting to Gazebo through PX4’s Gazebo transport interface and ROS~2 bridges.
Telemetry, control, and sensor data are synchronized using the Micro XRCE-DDS Agent, enabling bidirectional communication between PX4 and ROS~2 topics under a unified simulation clock.

\subsection{Localization}
Localization is managed by PX4’s onboard Extended Kalman Filter (EKF2), which fuses IMU measurements with external odometry supplied by Gazebo.
GPS input is disabled, and EKF2 receives continuous ground-truth pose updates equivalent to ideal visual–inertial odometry. This configuration ensures drift-free localization within a simulated GPS-denied environment.
Coordinate frames are consistently defined as:
\begin{itemize}
    \item \texttt{map}~$\rightarrow$~\texttt{odom}~$\rightarrow$~\texttt{base\_link} for global–local alignment, and
    \item \texttt{base\_link}~$\rightarrow$~\texttt{RGLLidar} for sensor placement.
\end{itemize}
Gazebo and mapping nodes operate in the East–North–Up (ENU) reference frame, whereas PX4 utilizes the North–East–Down (NED) convention.
A static transform aligns these frames using $(x_\text{NED}, y_\text{NED}, z_\text{NED}) = (y_\text{ENU}, x_\text{ENU}, -z_\text{ENU})$, thereby ensuring consistent interpretation of trajectories and waypoints by both systems.

\subsection{Perception and Mapping}
Perception is performed using the Robotec GPU LiDAR plugin, which generates physically accurate, ray-traced depth data accelerated by RTX GPUs.  
Each LiDAR publishes ROS~2 \texttt{LaserScan} messages at 10~Hz, representing dense range readings of the surrounding environment.

Mapping begins with a BIM-derived prior loaded through the \texttt{simple\_map\_publisher} node.  
The prior is stored as a binary \texttt{OccupancyGrid} (PGM+YAML) representing walls, corridors, and free regions.  
During operation, the \texttt{bim\_fusion\_node} fuses incoming LiDAR scans with this prior using a log-odds update model:
\begin{itemize}
    \item Cells hit by LiDAR endpoints increase occupancy log-odds.
    \item Cells along free-space rays decrease occupancy log-odds.
\end{itemize}
This process refines the BIM prior into an up-to-date, as-built occupancy map aligned with sensor observations.  
To maintain geometric stability, a $0.2 \mathrm{m}$ margin around BIM wall boundaries is frozen against log-odds updates, except at designated openings (e.g., doorways).  
The fusion node outputs:
\begin{enumerate}
    \item \texttt{/updated\_map} — fused occupancy probabilities,
    \item \texttt{/uncertainty\_map} — per-cell Shannon entropy, and
    \item \texttt{/discrepancy\_map} — absolute deviation from the BIM prior.
\end{enumerate}

\subsection{Path Planning and Global Navigation}
Global navigation is handled by the \texttt{bim\_global\_planner} node, which processes the fused occupancy map and computes feasible, collision-free trajectories using an RRT* planner.  
Obstacle inflation is applied according to the UGV’s radius to preserve safe clearances, and the resulting path is smoothed and resampled for continuous execution.  
Two synchronized outputs are published:
\begin{itemize}
    \item a \texttt{nav\_msgs/Path} for visualization in the map (ENU) frame, and
    \item a \texttt{Float32MultiArray} of \([N, E]\) waypoints for PX4, expressed in the NED frame.
\end{itemize}

\subsection{Control and Execution}
Low-level control for the UGV is implemented by the \texttt{rover\_waypoint\_navigator} node, which performs PX4 offboard velocity control.  
Velocity setpoints (\texttt{TrajectorySetpoint} messages) are streamed to \texttt{/fmu/in/trajectory\_setpoint} at 10~Hz, containing planar velocity components and a yaw command.  
The controller modulates speed adaptively:
\begin{itemize}
    \item reducing velocity under large yaw misalignment,
    \item accelerating when aligned with the waypoint direction, and
    \item issuing a stop command if the corridor becomes unsafe.
\end{itemize}
PX4’s \texttt{OffboardControlMode} ensures active offboard control, while \texttt{VehicleCommand} topics manage arming and mode transitions.  
Safety interlocks block motion until odometry and waypoint validation are confirmed.

\subsection{Active Sensing and Risk Assessment}
The \texttt{corridor\_assessor} node continuously evaluates navigation safety by computing the spatially averaged corridor risk $R_\ell(t)$, as defined in Eq.~\ref{func:risk}.  
Here, $r_t(i,j)$ represents the per-cell instantaneous risk composed of weighted uncertainty and discrepancy terms, while $R_\ell(t)$ denotes the mean risk within a look-ahead window $\mathcal{W}_\ell(t)$ along the UGV’s path.

In practice, the assessor subscribes to the fused occupancy, uncertainty, and discrepancy maps published by the \texttt{bim\_fusion\_node}.  
Within a configurable look-ahead distance (typically $5 \mathrm{m}$ ), it computes $R_\ell(t)$ online at each control step.  
If the aggregated corridor risk exceeds the predefined safety threshold $\tau_{\text{safe}}$, an \texttt{/unsafe} flag is broadcast, causing the UGV to halt motion.  
This event automatically triggers UAV intervention: the aerial robot performs an overhead re-scan of the high-risk region, updates the fused occupancy and uncertainty layers, and enables the ground robot to replan and continue navigation safely.

This module directly operationalizes the risk model introduced in Eq.~\ref{func:risk}, linking perception uncertainty and BIM discrepancies to real-time decision making within the active sensing framework.

\subsection{Multi-Robot Integration and Communication}
The UAV and UGV operate as independent PX4 instances within the same Gazebo world but share perception and mapping topics via ROS~2’s DDS transport.  
Simulation time is synchronized across all nodes to ensure deterministic message playback.  
The Micro XRCE-DDS Agent bridges PX4’s internal uORB middleware to ROS~2, enabling consistent telemetry and control exchange between vehicles.

\subsection{Compute Environment and Reproducibility}
All experiments are performed on Ubuntu~22.04 with an NVIDIA RTX-series GPU and CUDA acceleration.  
LiDAR simulation, map fusion, and motion planning run in real time.  
Map resolution (0.1 $\mathrm{m}$), LiDAR field of view ($360^{\circ}$), robot radius, and corridor look-ahead distance are kept constant across trials for reproducibility.  
All ROS~2 messages are time-stamped using simulation time to guarantee deterministic replay.

\subsection{Summary}
In summary, the proposed system integrates:
\begin{enumerate}
    \item PX4 for flight and ground control,
    \item Gazebo Harmonic for physics-based simulation,
    \item Robotec GPU LiDAR for high-fidelity sensing,
    \item BIM-based probabilistic fusion for dynamic map refinement,
    \item RRT*-based path planning, and
    \item PX4 offboard velocity control with real-time risk assessment.
\end{enumerate}
Together, these modules create a reproducible UAV–UGV simulation and navigation framework capable of performing autonomous mapping, path planning, and safety-aware decision making in BIM-referenced environments.

\section{Experiments and Results}
\label{sec:experiments}
\subsection{Setup}
\begin{figure}[H]
    \centering
    \includegraphics[width=\columnwidth,keepaspectratio]{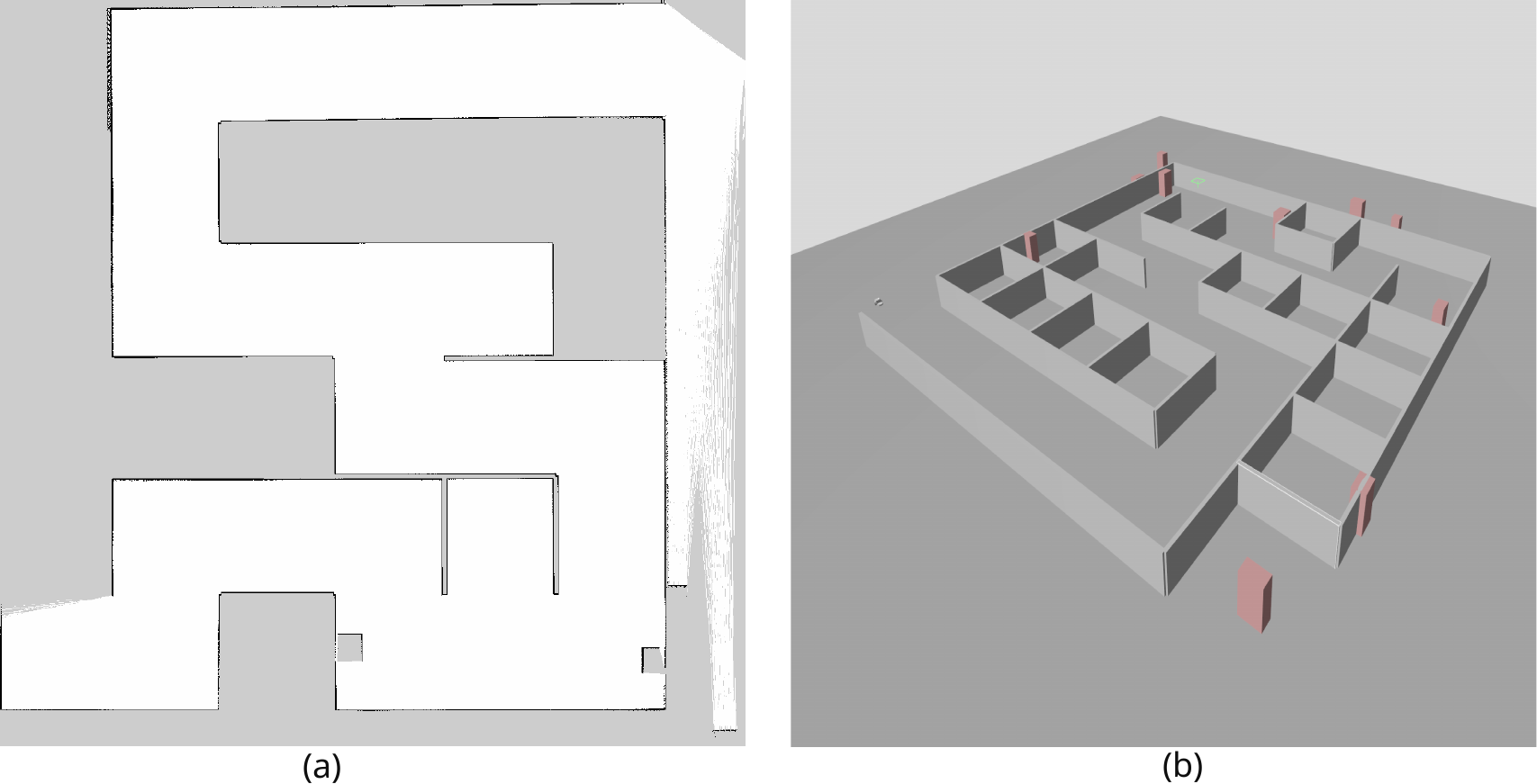}
    \caption{
    Simulation environment and BIM-based prior map. 
    (a) Gazebo simulation environment representing the construction scene used for UAV--UGV experiments. 
    (b) Corresponding 2D occupancy map generated from BIM.
    }
    \label{fig:gazebo_bim}
\end{figure}

We conducted all experiments in the Gazebo Harmonic simulator (see \cref{fig:gazebo_bim}(a)), using PX4--ROS 2 integration and RGL.
Our UAV and UGV operated inside a construction environment modeled from BIM as an indoor corridor network ($50\ \mathrm{m} \times 50\ \mathrm{m}$). 
We converted the BIM prior into a 0.1\,m-resolution occupancy grid map ($500 \times 500$ cells) (see \cref{fig:gazebo_bim}(b)). 
The UGV followed an RRT* path between fixed start and goal points, moving at a steady speed of 1\,m/s. 
Whenever needed, the UAV performed lateral sweeps at a height of 2.5\,m.

\subsection{Experimental scenarios}
We evaluated our approach across four representative site conditions:

\begin{enumerate}
    \item \textbf{Scenario~1: BIM Mismatch (Static Prior Failure).}  
    In this baseline experiment, the UGV navigated solely using the BIM-derived occupancy map, with no aerial assistance or online updates.  
    A new wall was introduced into the Gazebo environment but was not represented in the BIM prior (\cref{fig:static-bim-map}).  
    After traveling approximately $70\,\mathrm{m}$, the UGV’s LiDAR detected a rapid increase in corridor risk $(R_\ell = 0.69)$, indicating an unexpected obstruction.  
    Since the global planner continued to rely on the outdated prior, the planned trajectory intersected the newly added wall, causing the UGV to halt and terminate the mission.   
    This scenario highlights the inherent limitations of static BIM priors and motivates the need for adaptive, perception-driven map correction mechanisms.

    \begin{figure}[H]
    \centering
    \includegraphics[width=\columnwidth,keepaspectratio]{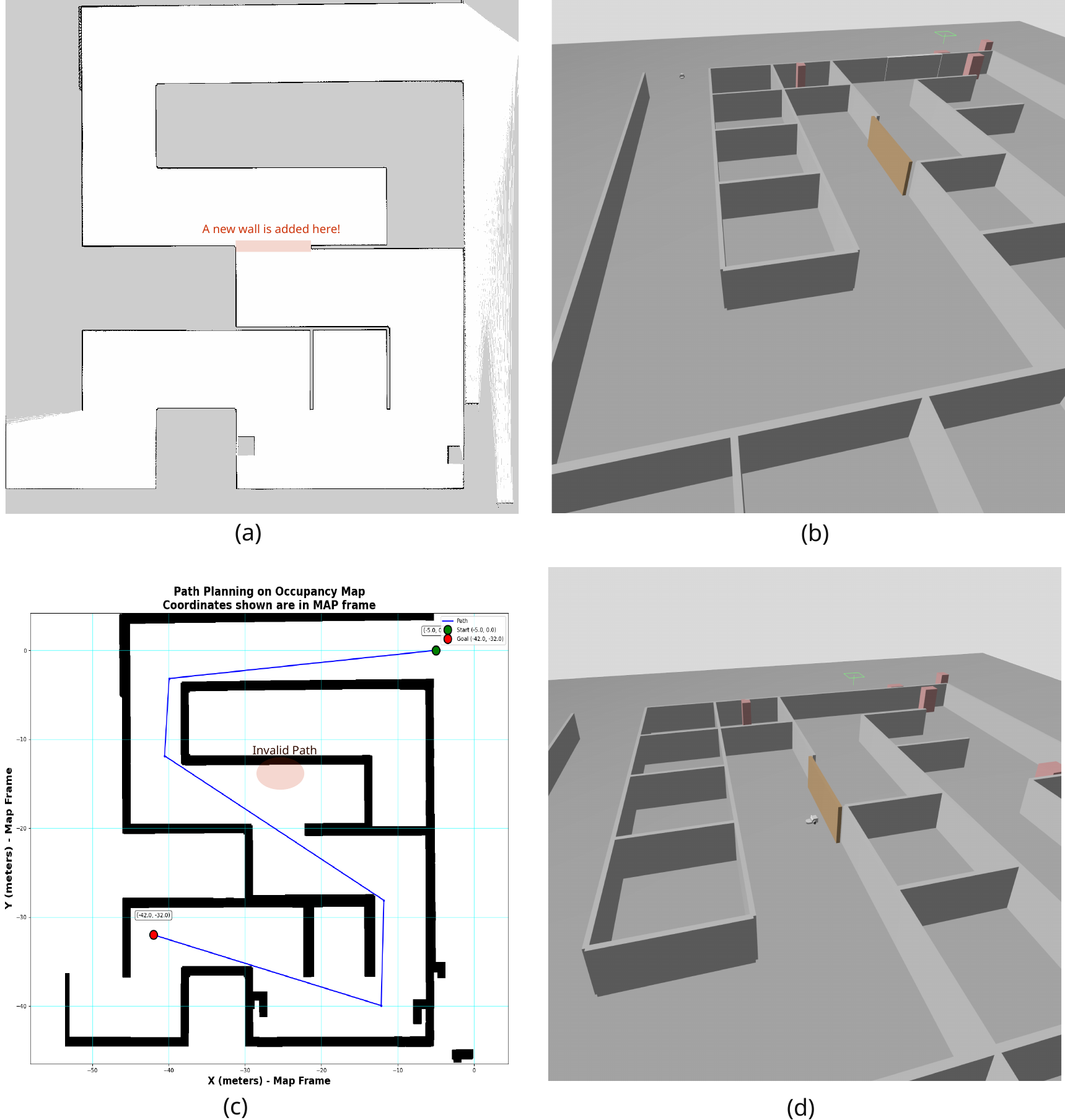}
    \caption{
    \textbf{Scenario~1 – BIM mismatch under static prior (UGV halt).}
    (a) BIM-based prior occupancy map where a new wall has been introduced on site but is absent from the BIM prior.  
    (b) Gazebo simulation environment showing the updated scene with the newly inserted wall (highlighted in red).  
    (c) Global path planned on the outdated occupancy map, producing an invalid route that intersects the new obstacle.  
    (d) UGV halts operation after detecting reduced clearance near the unmodeled wall, demonstrating the limitation of static BIM-based navigation without active updates.
    }
    \label{fig:static-bim-map}
\end{figure}

    \item \textbf{Scenario~2: BIM Mismatch Correction (UAV-Assisted Update).}  
    Expanding on the previous experiment, we introduced several intentional discrepancies between the BIM model and the as-built environment—specifically, by adding two new walls and removing one existing wall.
    As the UGV navigated its predetermined path based on the static BIM prior, it encountered an unforeseen obstruction. This resulted in a marked increase in corridor risk ($R_\ell = 0.69$) after approximately $70\,\mathrm{m}$ of travel.
    After the risk exceeded the predefined safety threshold ($\tau_{\text{safe}}$), the UGV automatically halted and initiated a UAV re-scan of the affected area (\cref{fig:static-bim-map-recan}).
    The UAV conducted an overhead sweep using \texttt{slam\_toolbox} to update both occupancy and uncertainty maps, thereby correcting structural inconsistencies, including missing or displaced walls.
    As illustrated in \cref{fig:scenario1_rrescan}, the corridor risk decreased rapidly following the map update, falling below the safety threshold. The UGV then successfully replanned its path using the corrected occupancy grid.
    The new trajectory circumvented the previously obstructed corridor, enabling the robot to reach its target. The mission was completed with a total travel distance of $109.2\,\mathrm{m}$ in $198.6\,\mathrm{s}$.
    This scenario demonstrates the framework’s capability for autonomous detection and correction of BIM discrepancies through UAV-initiated map updates, thereby restoring safe and uninterrupted navigation.

    \begin{figure}[H]
    \centering
\includegraphics[width=\columnwidth,keepaspectratio]{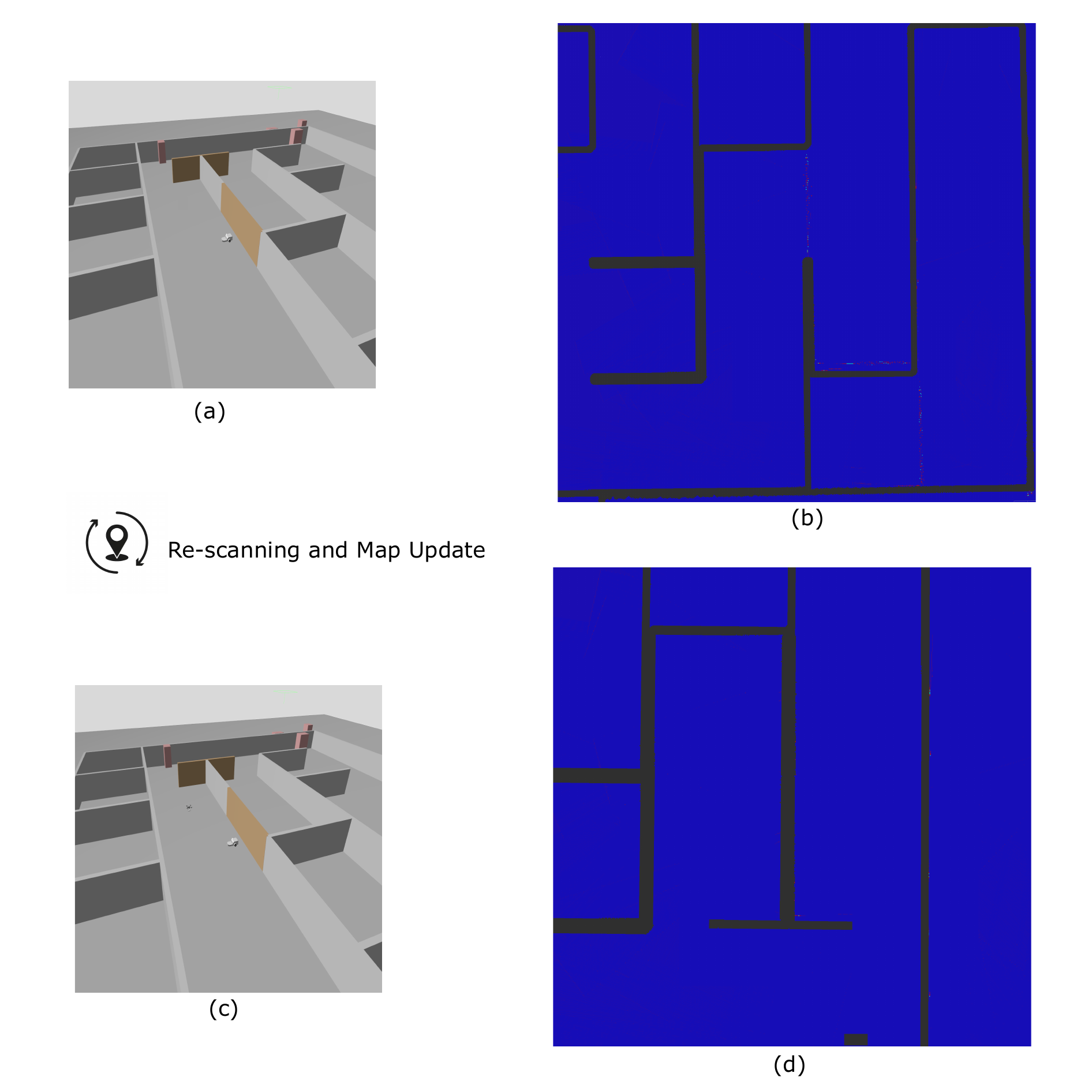} 
    \caption{
    \textbf{Scenario~2 – UAV-triggered re-scanning and map update.}
    (a, b) The UGV halts after detecting a high corridor risk due to unmodeled structural changes—two new walls added and one removed, creating elevated uncertainty in the costmap (represented by lighter regions).  
    (c, d) Following the triggered UAV re-scan, the updated occupancy and uncertainty maps reflect corrected geometry, where the previously ambiguous regions have been resolved.  
    In (b) and (d), the occupancy costmap is visualized using the Gazebo convention: dark blue cells correspond to free (non-occupied) space. At the same time, gray regions indicate occupied cells such as walls and obstacles.
    This process reduces overall entropy and enables safe replanning toward the goal.
    }
    \label{fig:static-bim-map-recan}
\end{figure}

\begin{figure}[H]
    \centering
    \includegraphics[width=\columnwidth,keepaspectratio]{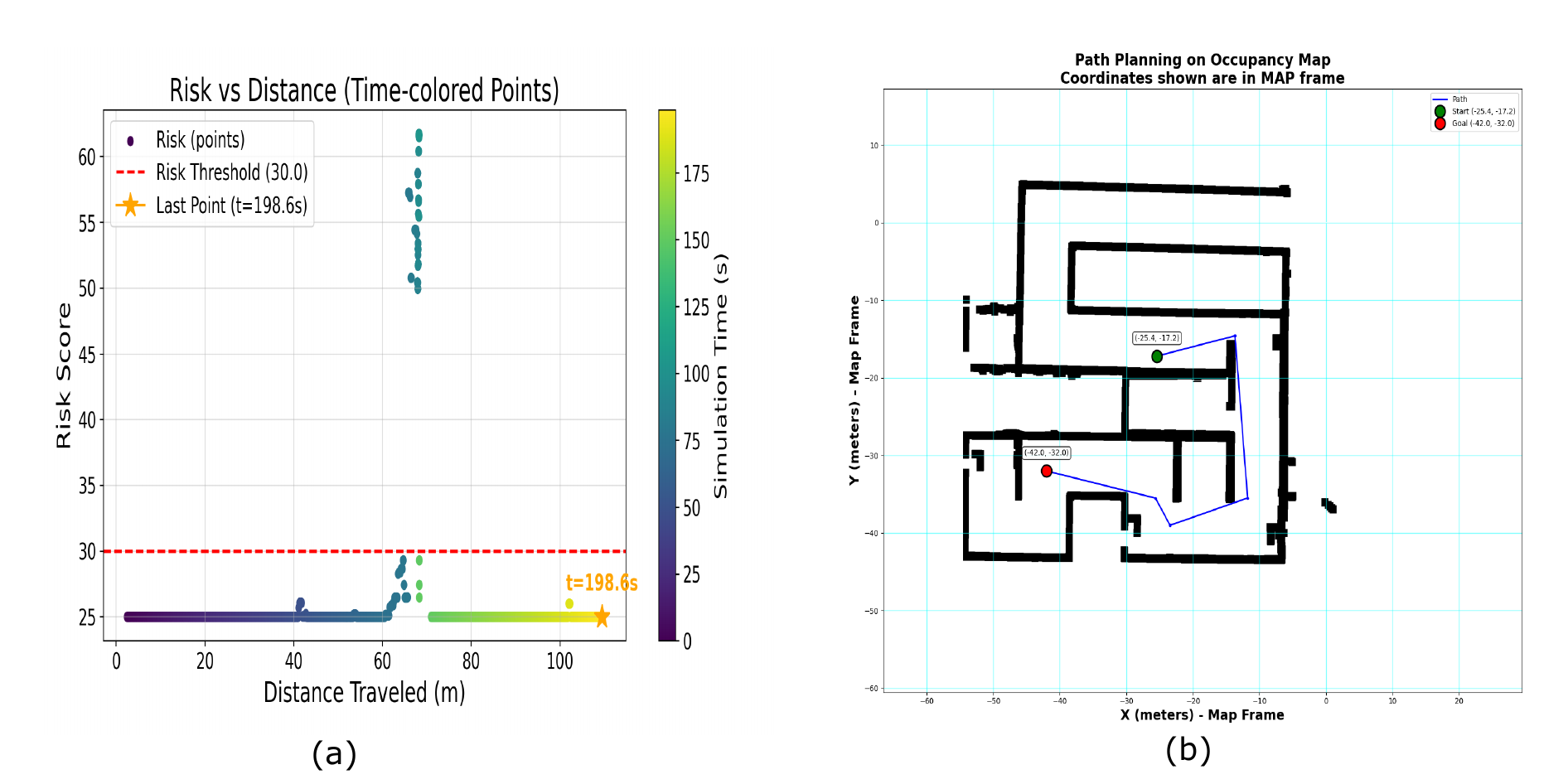} 
    \caption{
    \textbf{Scenario~2 – Risk evolution and replanning after UAV re-scan.}  
    (a) Risk score as a function of distance traveled. The corridor risk $R_\ell$ rises sharply upon encountering unmodeled structural changes, surpassing the safety threshold (red dashed line) and triggering a UAV re-scan. After the UAV updates the occupancy map, the risk level drops below the threshold, allowing the UGV to resume navigation.  
    (b) Updated global path on the corrected occupancy map. The replanned trajectory successfully avoids the newly added walls and guides the UGV to the goal, demonstrating safe and adaptive navigation following autonomous map correction.
    }
    \label{fig:scenario1_rrescan}
\end{figure}

        \item \textbf{Scenario 3: Occlusion Recovery via UAV-Assisted Map Update.}
    In this experiment, the BIM prior accurately reflected the environment; however, temporary clutter was introduced by placing six stacked boxes along the corridor.
    Although these boxes did not physically obstruct the planned path, they partially occluded the UGV LiDAR field of view. This partial occlusion resulted in incomplete local mapping and increased uncertainty in the occupancy grid.
    As illustrated in \cref{fig:occlusion-recovery}(a–b), this occlusion produced brighter regions of uncertainty in the costmap and caused the corridor risk to exceed the safety threshold ($\tau_{\text{safe}}$), prompting the UGV to halt.
    Once stationary, the UGV autonomously triggered a UAV re-scan of the affected region.
    The UAV conducted an overhead sweep, capturing the occluded area and updating both the occupancy and uncertainty layers to accurately reveal the free-space geometry (\cref{fig:occlusion-recovery}(c–d)).
    This re-scanning process significantly reduced map entropy and restored reliable perception for subsequent path planning.
    As depicted in \cref{fig:occlusion-risk}, the risk score rose sharply at approximately $40\,\mathrm{m}$ of travel prior to UAV intervention, but dropped below the threshold immediately after the updated map was integrated.
    Following this correction, the UGV resumed navigation and successfully reached its goal, completing the mission in $201.1\,\mathrm{s}$.
    This scenario demonstrates the system’s capability to autonomously detect and resolve perception degradation caused by temporary occlusions, thereby ensuring robust and uninterrupted operation even when the BIM prior remains structurally accurate.

    \begin{figure}[H]
    \centering
    \includegraphics[width=\columnwidth,keepaspectratio]{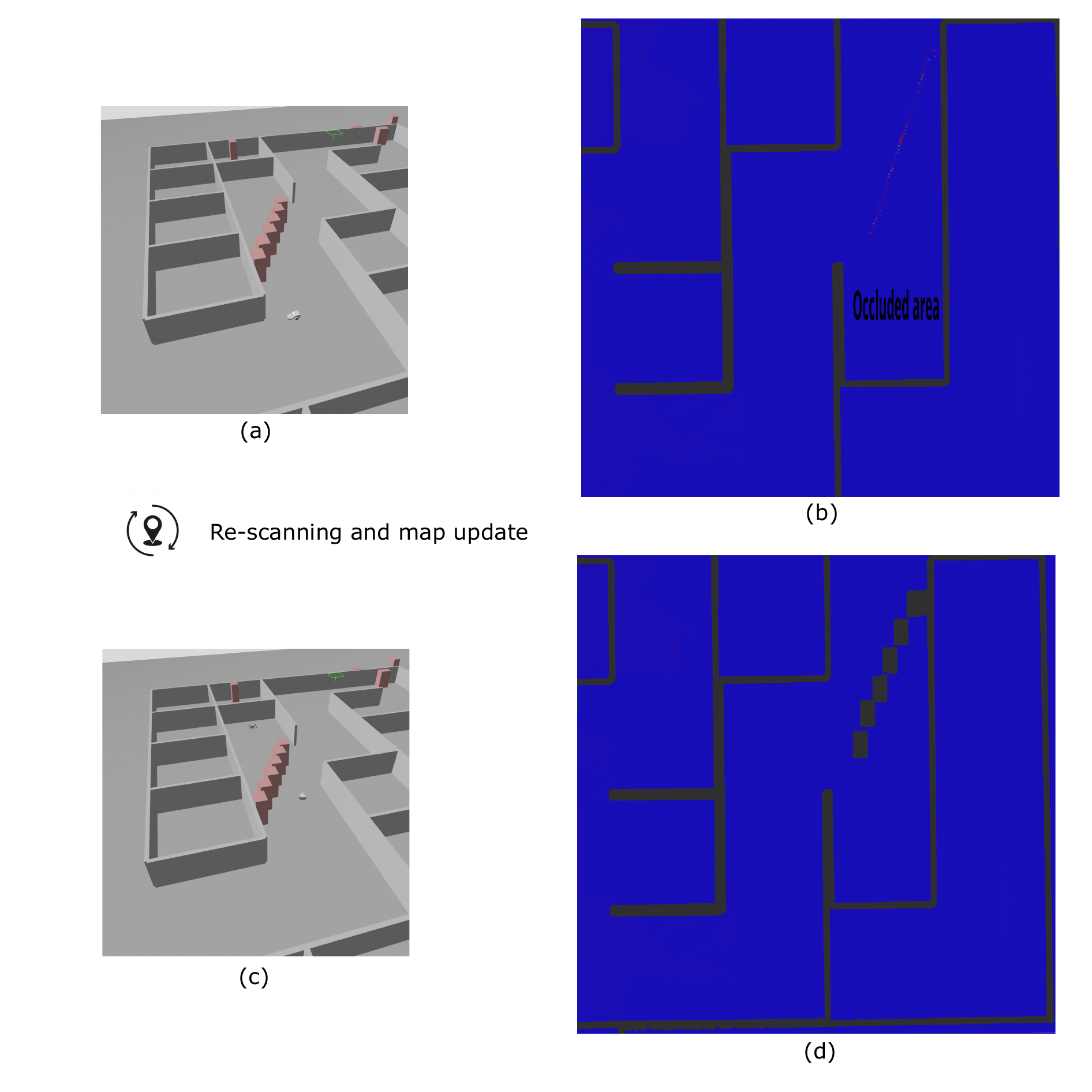} 
    \caption{
    \textbf{Scenario~3 – UAV-Assisted Occlusion Recovery and Map Update.}
    (a, b) The UGV encounters temporary clutter in the form of six stacked boxes, which partially
    occlude its LiDAR field of view. This occlusion increases local uncertainty (indicated by lighter
    regions) in the occupancy costmap. In (b), the occupancy map follows the Gazebo costmap
    color convention: dark blue represents free (non-occupied) space, while gray cells denote occupied
    regions such as walls and obstacles. Although the stacked boxes do not physically block the planned
    corridor, the loss of visibility elevates the corridor risk above the safety threshold, causing the UGV
    to halt. 
    (c, d) Following the triggered UAV overhead re-scan, the updated occupancy and uncertainty maps
    accurately reconstruct the previously occluded free-space region (again shown in dark blue), reducing
    entropy and restoring a reliable understanding of the environment. This UAV intervention resolves
    the ambiguity caused by occlusion and enables the UGV to safely resume navigation.
    }
    \label{fig:occlusion-recovery}
\end{figure}

\begin{figure}[H]
    \centering
    \includegraphics[width=0.5\columnwidth,keepaspectratio]{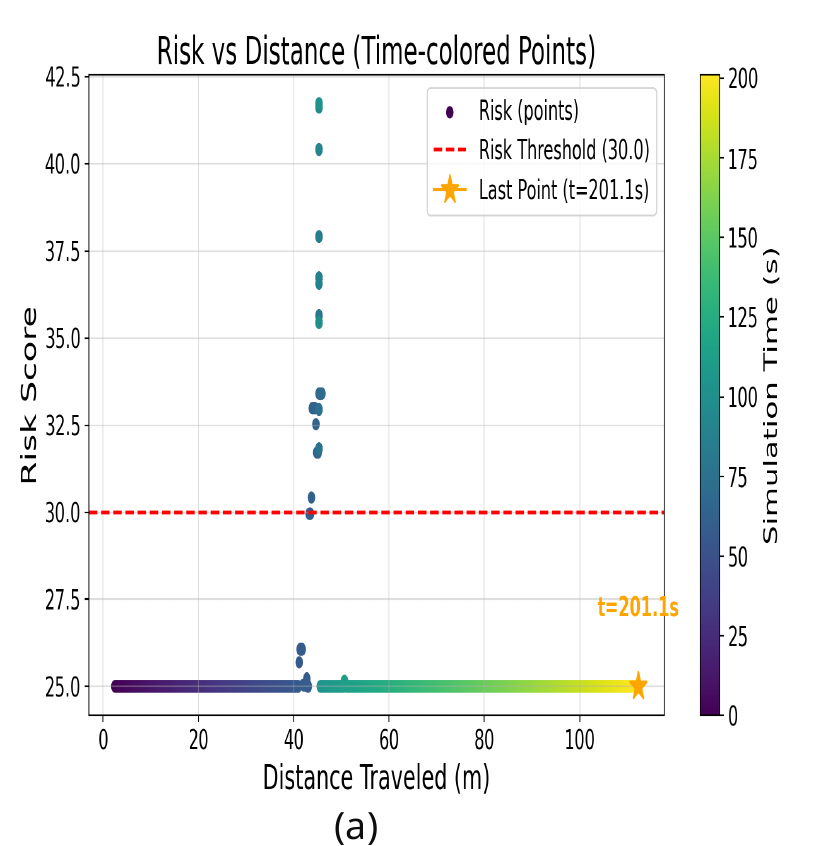} 
    \caption{
    \textbf{Scenario 3 – Risk Evolution During Occlusion Recovery.}
    The corridor risk rises sharply when the UGV encounters a region partially occluded by stacked boxes. This increase surpasses the safety threshold (indicated by the red dashed line), prompting a UAV-initiated re-scan of the affected area.
    Following the UAV’s update of the occupancy and uncertainty maps, the perceived risk drops below the threshold. This enables the UGV to safely resume navigation and complete the mission in $201.1\,\mathrm{s}$.
    }
    \label{fig:occlusion-risk}
\end{figure}

    \item \textbf{Scenario 4: Frontier-Based Exploration (UGV-Only).}
    In this experiment, the environment was configured identically to Scenario 2, with the addition of two new walls and the removal of one existing wall, thereby introducing BIM–as-built discrepancies along the corridor.
    Rather than deploying the UAV, the UGV relied exclusively on frontier-based exploration to address the perceived inconsistencies.
    Upon detecting elevated corridor risk, the robot halted and initiated local exploration within a 5-meter radius, autonomously selecting frontier cells and navigating toward them to gather additional LiDAR data.
    This process gradually reduced local uncertainty and enabled the robot to reconstruct the corrected occupancy structure without external assistance.
    However, since the UGV was required to physically traverse each frontier cell to obtain new observations, the entire process required nearly twice the mission duration compared to the UAV-assisted approach.
    Although the final levels of uncertainty and risk were comparable to those achieved with UAV re-scanning, the recovery process was substantially slower and less efficient. We omit visual figures for Scenario 4 because its qualitative map updates, occupancy corrections, and replanning behavior are nearly identical to Scenario 2; the main difference lies in the longer mission duration resulting from UGV-only exploration. This outcome underscores the advantage of overhead sensing for rapid map correction.
    \end{enumerate}
    
\begin{table}[t]
\centering
\caption{Percent improvements and mission outcomes across Scenarios~1--4 (means over five runs). Risk and entropy reductions are computed \emph{at the first intervention} as $\Delta R_{\%} = 100\,\frac{R_\text{before}-R_\text{after}}{R_\text{before}}$ and $\Delta H_{\%} = 100\,\frac{H_\text{before}-H_\text{after}}{H_\text{before}}$, where "before" is measured immediately before the UAV re-scan or frontier routine, and "after" immediately following map fusion and re-planning. "Min.~Clr." is the minimum EDT clearance ($\mathrm{m}$) along the executed path (post inflation), "Path" is the total path length ($\mathrm{m}$), and "Time" is the total mission duration ($\mathrm{s}$).}
\label{tab:s1s4}
\resizebox{0.9\columnwidth}{!}{%
\begin{tabular}{lccccc c}
\toprule
\textbf{Scenario} & $\mathbf{\Delta R_{\%}}$ $\uparrow$ & $\mathbf{\Delta H_{\%}}$ $\uparrow$ & \textbf{Min.~Clr. [m]} $\uparrow$ & \textbf{Path [m]} $\downarrow$ & \textbf{Time [s]} $\downarrow$ & \textbf{Goal} \\
\midrule
\textbf{1} & 0 & 0 & 0.20 & 70.0 & 104.0 & $\times$ \\
\textbf{2} & \textbf{58} & \textbf{43} & \textbf{0.45} & 109.2 & 198.6 & $\checkmark$ \\
\textbf{3} & 40 & 41 & 0.42 & \textbf{108.2} & 201.1 & $\checkmark$ \\
\textbf{4} & 55 & 39 & 0.44 & 118.3 & 351.2 & $\checkmark$ \\
\bottomrule
\end{tabular}
}
\end{table}

In all our experiments, our risk-triggered UAV re-scanning framework consistently delivered the best results for safety and uncertainty reduction, with minimal impact on mission time.
As shown in \cref{tab:s1s4}, Scenario 2 resulted in a 58\% reduction in corridor risk and a 43\% decrease in entropy. This demonstrates that the system can automatically spot BIM–as-built mismatches and restore safe navigation.
In Scenario 3, even when there was significant occlusion and clutter, using the UAV to update the map reduced risk by 40\% and entropy by 41\%. This confirms that the approach is robust even when sensor perception is challenged.
The frontier-based baseline in Scenario 4 eventually achieved a similar level of uncertainty reduction, but it took nearly twice as long. This highlights how much more efficient aerial re-scanning can be.
Overall, our results show that using UAV interventions when needed helps keep risk low and the map reliable, while also avoiding extra exploration time and computation.

\section{Conclusion}
\label{sec:conclusion}

This paper presented a \emph{BIM-discrepancy-driven active sensing framework} for cooperative UAV and UGV navigation in dynamic construction environments.
By leveraging BIM as probabilistic priors and continuously updating them with real-time LiDAR observations, the system enables adaptive and risk-aware mapping, eliminating the need for continuous exploration or dense 3D reconstruction.
A corridor-based risk formulation integrates occupancy uncertainty, BIM–map discrepancy, and clearance, thereby providing a unified criterion for triggering UAV rescans when navigation safety is compromised.
Through PX4–Gazebo simulations utilizing RGL and ROS~2, the framework demonstrated the ability to detect structural changes, reduce mapping uncertainty, and maintain safe UGV operation in evolving site conditions.

Experimental results verified that UAV-assisted updates significantly reduced mean corridor risk and improved minimum clearance compared to navigation using only BIM.
In particular, the framework autonomously corrected geometric mismatches between the BIM and the as-built environment and effectively recovered from occlusions that degraded UGV perception.

\textbf{Limitations and Future Work.}
Several specific limitations warrant discussion. First, our 2D occupancy representation assumes vertical homogeneity within the scanner's elevation range. In environments with overhanging obstacles (e.g., partially constructed floors, suspended equipment), this assumption may lead to unsafe clearance estimates. Future work will extend the system toward 2.5D and volumetric mapping, for example, using OctoMap or Voxblox, to incorporate vertical clearance estimation and reasoning about overhanging obstacles.

Second, the static risk threshold requires manual tuning based on site-specific safety requirements and operational constraints. Adaptive threshold selection based on historical corridor statistics could improve robustness. We plan to incorporate \emph{semantic-aware risk assessment}, in which both BIM-derived object classes (such as structural walls, equipment, and scaffolds) and onboard semantic perception inform risk computation and UAV rescan priorities. This semantic context will enable the system to reason about the functional importance and dynamic behavior of site elements, rather than relying solely on geometric information.

Third, our evaluation in simulation does not capture real-world sensor degradation from dust, lighting variations, or GPS-denied localization drift. While the simulation-based validation provides a controlled and reproducible environment, future work will validate the framework on physical hardware to account for real-world sensor noise and environmental factors. 

We also envision extending this framework toward \emph{V2X-enabled multi-robot coordination}, in which multiple ground and aerial robots share real-time risk, map, and intent information over local communication networks to maintain situational awareness across large-scale sites collaboratively. The proposed framework demonstrates that BIM-informed and risk-triggered sensing provides a scalable pathway toward autonomous, safety-conscious navigation in dynamic construction environments. This approach bridges the gap between design-time information and real-time field operations. Integrating BIM semantics, active sensing, and connected autonomy can ultimately enable predictive, cooperative, and safety-aware perception for construction robotics at scale.

\section*{Acknowledgements}
The presented work has been supported by the U.S. National Science Foundation (NSF) CAREER Award through grant No. CMMI 2047138. The authors gratefully acknowledge the support from the NSF. Any opinions, findings, conclusions, and recommendations expressed in this paper are those of the authors and do not necessarily represent those of the NSF.

\bibliographystyle{elsarticle-num}
\bibliography{references}

\end{document}